# Some properties of the Ukrainian writing system

*Solomija Buk*[1]*, Lviv*
*Ján Mačutek*[2]*, Bratislava*
*Andrij Rovenchak*[3]*, Lviv*

**Abstract.** We investigate the grapheme–phoneme relation in Ukrainian and some properties of the Ukrainian version of the Cyrillic alphabet.

## 1. Introductory remarks

Ukrainian is an East Slavic language spoken by about 40 million people in Ukraine and Ukrainian communities in neighboring states (Belarus, Moldova, Poland, Slovakia, Russia — especially in the so-called *Zelenyj Klyn* 'Green wedge' in the Far East Siberia from the Amur and Ussuri rivers eastwards to the Pacific), also in Argentina, Australia, Brazil, Canada, USA, and some others.

The features typical for modern Ukrainian are found already in the texts from 11th-12th cent. AD, they have been appearing systematically since 14th-15th cent. (Rusanivsjkyj 2004). Ukrainian uses the Cyrillic script. The Cyrillic alphabet, also known as *azbuka* (from old names of its first two letters А (азъ) and Б (боукы)), has been traditionally used to write East and South Slavic languages (with the exception of modern Croatian and Slovenian), and also Romanian until 1860 (Jensen 1969: 491). As a result of political decisions it spread over a much larger area covering most (but not all) of languages in the former USSR, many of them using Latin or Arabic script before (cf. Comrie 1996b for a more detailed historical overview). Obviously, being applied in so different languages like Russian, Abkhaz, Tatar, Tajik or Chukchi (to give just a few examples) it had to represent much more phonemes than those occurring in Slavic languages, hence there are/were many language specific modifications of the alphabet (modified particular letters, diacritic marks or completely new letters, cf. Comrie 1996a). The Ukrainian version of the Cyrillic alphabet is called also *abetka* in vernacular from the names of the first two letters *a* and *be*. It consists of 33 letters:

< А а, Б б, В в, Г г, Ґ ґ, Д д, Е е, Є є, Ж ж, З з, И и, І і, Ї ї, Й й, К к, Л л, М м, Н н, О о, П п, Р р, С с, Т т, У у, Ф ф, Х х, Ц ц, Ч ч, Ш ш, Щ щ, ь, Ю ю, Я я >

[1] Department for General Linguistics, Ivan Franko National University of Lviv, 1 Universytetska St., Lviv, UA-79000, Ukraine, e-mail: solomija@gmail.com.
[2] Department of Applied Mathematics and Statistics, Comenius University, Mlynská dolina, 842 48 Bratislava, Slovakia, e-mail: jmacutek@yahoo.com.
[3] Department for Theoretical Physics, Ivan Franko National University of Lviv, 12 Drahomanov St., Lviv, UA-79005, Ukraine, e-mail: andrij@ktf.franko.lviv.ua, andrij.rovenchak@gmail.com.

When italicized, the following lowercase letters differ more or less significantly from the roman type: г — *г,* д — *д,* и — *и,* й — *й,* п — *п,* т — *т,* ш — *ш.*

Two letters are usually considered unique in the Ukrainian alphabet: < Ґ > and < Ї >. The first one denotes velar plosive [g] and is used mainly in loan-words. This letter was first attested in 16th cent. and included in the alphabet by Meletius Smotrytsky in 1619 (Pivtorak 2004a). The use of < Ґ > was abolished by Stalin's regime in 1933 and reintroduced into the Ukrainian alphabet only in 1990. Regarding the uniqueness of this letter one must note however that < Ґ > was in use in the Belarusian orthography before 1933 but never officially revived until today except in some dissident editions (Katkouski and Rrapo n.d.), cf. also Barry 1997. The letter < Ї > in its modern phonetic value ([ji], see details below) was first attested in 1875 (Pivtorak 2004b) becoming thus the last standardized letter of the Ukrainian alphabet.

The apostrophe < ' > is not considered as a part of the alphabet but it plays an important role in the orthography (similar to that of the hard sign < ъ > in Russian), as described below. Ukrainian orthography is largely phonemic and thus can be referred to as a 'shallow' one (Coulmas 2004: 380). The deviations from the 'one letter to one sound' correspondence are few and the sound changes due to the assimilation are quite predictable and justified on the morphological level.

The letter < щ > always represents a two-phoneme combination /ʃʧ/ = /ʃ/ + /ʧ/.

The letter < ь > ('soft sign') is not given in a capital form, as it never stands in a word-initial position. This letter does not represent any sound but indicates the palatalization of a preceding consonant. The letters < є, ю, я > can represent one or two phonemes, depending on their position. When immediately following a consonant, they indicate the palatalization of the consonant and correspond to /ɛ, u, ɑ/, respectively. In a word-initial position, after < а, е, и, і, о, у, є, ї, ю, я >, < ь > and the apostrophe < ' > these letters represent two-phoneme combinations with /j/: /jɛ, ju, jɑ/. In modern Ukrainian the letter < ї > always corresponds to /ji/, nevertheless, it must be separated by the apostrophe from the preceding consonant. Historically, this letter originated from older < ѣ > and had a two-fold correspondence similar to < є, ю, я >, both /i/ with palatalizing a preceding consonant and /ji/. However, in modern literary Ukrainian the letter < і > replaced < ї > in the first case.

Originally, the palatalization of a preceding (mainly dental) consonant by < і > did not occur if this letter originated from older < о > (this fact is seen in < о > preserved in some word-forms: *стіл / стола, дім / дому, ніс / носа* versus *ніс / нести*). At present, such pronunciation, being influenced by the orthography, gradually becomes marginal though it is not considered incorrect.

Note that there is no special letter to indicate the palatalization before /ɔ/, unlike Russian or Belarusian < ё >. In Ukrainian, the combination < ьо > is used in this case.

We would like to note that in Ukrainian the letter < і > is used to represent the phoneme /i/. Within Slavic languages using the Cyrillic script only Belarusian has the same practice, in all the other orthographies the letter < и > is used to represent this phoneme. In Ukrainian, however, the grapheme < и > corresponds to the phoneme /ɪ/ (not to be mixed with close-central /ɨ/ typical for, e.g., Polish and Russian). Another difference with East Slavic orthographies is that the use of < е > is consistent with Slavic Latin and South Slavic Cyrillic practice. A special grapheme < є >, which inherited its outer form from old Cyrillic alphabet, has a two-fold nature serving to denote both the palatalization of a preceding consonant + /ɛ/ and /jɛ/, but not < e > — as in Russian and Belarusian.

## 2. Ukrainian phonetics

### 2.1. General

There is no universally accepted definition for the notion of phoneme in scientific literature. In this work, we consider a phoneme as a group of phonetically similar speech sounds. It is the smallest structural unit of language that can distinguish meaning of the words. This definition is close to, e. g., the Saint-Petersburg (Leningrad) School of phonology or American descriptivism versus, e. g., Moscow School of phonology. In particular, we consider the assimilatory changes within one morpheme as different phonemes but not as allophonic modifications. Our approach is consistent with the similar existing studies on the Slavic languages: Slovak (Nemcová and Altmann 2008) and Slovene (Kelih 2008).

It is commonly accepted that Ukrainian has 38 phonemes: 6 vowels and 32 consonants (Bilodid 1969; Ponomariv 2001; Žovtobrjukh and Khomenko 2004). The deviations from this number are linked with different approaches to the phonemic status of semi-palatalized and geminate consonants (Žovtobrjukh and Kulyk 1965: 109–110). The details of pronunciation further are given mainly according to Pohribnyj 1984. The IPA transcription is based on the tables given by Bilous (2005). We would also like to note an English-language source for Ukrainian phonetics (Zilynśkyj 1979).

### 2.2. Vowel phonemes

The vowel phonemes are /ɑ, ɛ, ɪ, i, ɔ, u/. For Ukrainian vowels, the difference between stressed and unstressed positions is not crucial. When unstressed, /ɑ/ has an allophone [ɐ], /ɔ/ has an allophone [o], this sound also slightly approaches /u/ if followed by a syllable containing /u/ or /i/, /u/ has an allophone [ʊ], the variations in the pronunciation of /i/ are very slight. Most problems concern the difference between unstressed /ɛ/ and /ɪ/. Depending on the phonetic environment, several variations of these sounds can be identified. In Ukrainian phonetic transcription based on the Cyrillic script they are denoted as [e$^{и}$] (closer to [ɛ]) and [и$^{e}$] (closer to [ɪ]). We will join them in one allophone [e] belonging to the phoneme /ɛ/ as it seems incorrect — within our approach — to relate one allophone with different phonemes.

### 2.3. Consonant phonemes

Consonants in Ukrainian appear, along with ordinary ('hard') forms, in palatalized ('soft') or semi-palatalized ('semi-soft') variants. The first group consists of 22 phonemes: /b, ʋ, ɦ, g, d̪, ʒ, z̪, k, l̪, m, n̪, p, r, s̪, t̪, f, x, ʦ̪, ʧ, ʃ, ʣ̪, ʤ/. In the following text, we will not mark the dental character of the phonemes for simplicity.

The group of palatalized consonants consists of 10 phonemes: /j, dʲ, zʲ, lʲ, nʲ, rʲ, sʲ, tʲ, ʦʲ, ʣʲ/. There is no complete agreement about the nature of the palatalization of

/rʲ/, sometimes it is considered as a semi-palatalized consonant (Ponomariv 2001: 16, 20). As there is no special IPA mark for semi-palatalization, we will use a superscript dotless ‘j’, e. g., /r$^ȷ$/. The palatalization of the consonants /b$^ȷ$, ʋ$^ȷ$, ɦ$^ȷ$, ɡ$^ȷ$, ʒ$^ȷ$, k$^ȷ$, m$^ȷ$, p$^ȷ$, f$^ȷ$, x$^ȷ$, tʃ$^ȷ$, ʃ$^ȷ$, dʒ$^ȷ$/ is even weaker; they are usually treated rather as the allophones of the respective ‘hard’ consonants, not as separate phonemes.

Ukrainian has the following sonorants: /ʋ, l, lʲ, m, nʲ, r, r$^ȷ$, j/. The labio-dental approximant /ʋ/ represented by < в > must not be mixed with, e. g., Polish or Russian fricative /v/, which falls into a pair with voiceless /f/. In Ukrainian, fricative /f/ is quite rare phoneme appearing only in loans and in onomatopoetic words. The Ukrainian phoneme /ʋ/ can appear in several allophonic modifications:

- non-syllabic [u] — [u̯] starts a syllable coda (*мав, був, мавпа, шовк*), in continuous speech this sound can be found in a word-initial position after a vowel of the preceding word (*а вперше* [ɐu̯pɛrʃe]).
- voiced labialized velar approximant [w] before /ɔ, u/ and voiced consonants (not after a vowel): *вниз, вона, вухо*;
- voiceless labialized velar approximant [ʍ] before voiceless consonants (not after a vowel): *вперше* [ʍpɛrʃe];
- semipalatalized labio-dental approximant [ʋ$^ȷ$]: *він* [ʋ$^ȷ$in], *свято*[sʲʋ$^ȷ$ɑto].

In a syllable-final position (being more precise, the first position of a syllable coda) the phoneme /j/ represented by < й > appears as a non-syllabic sound [ i̯ ]: *хай, знайте*.

Sometimes the combinations of a vowel plus non-syllabic [u̯] or [ i̯ ] are considered as diphthongs ([ɑu̯], [uu̯], [ɔi̯ ], etc.) but they are not phonemic in Ukrainian.

Most obstruents can be grouped into the “voiced–voiceless” pairs: /b/–/p/, /ɡ/–/k/, /d/–/t/, /ʒ/–/ʃ/, /z/–/s/, /ʣ/–/ʦ/, /ʤ/–/ʧ/, /dʲ/–/tʲ/, /zʲ/–/sʲ/, /ʣʲ/–/ʦʲ/.

The articulation of the sound represented by < г > as voiced velar fricative [ɣ] instead of [ɦ] is incorrect. That is, the opposition between < г > and < х > (phonetically /ɦ/ and /x/) is not exact. Voiceless /f/ has no voiced counterpart.

No separate letters exist for the phonemes /ʣ/ and /ʤ/. They are represented by digraphs <дз> and <дж>, respectively: *дзвоник* /ʣʋɔnɪk/, *бджола* /bʤɔlɑ/. On the prefix-root boundary, however, these digraphs represent two phonemes: *надзвичайно* /nɑdzʋɪʧɑjnɔ/ (assimilates to /nɑʣzʋɪʧɑjnɔ/).

Also, no separate graphemes exist for the palatalized phonemes. To represent this feature, several techniques are used, see Table 2.

### 2.4. On the phonemic status of semi-palatalized consonants

As it was mentioned above, the following consonants have a semi-palatalized form: labials /b$^ȷ$, ʋ$^ȷ$, m$^ȷ$, p$^ȷ$, f$^ȷ$/, velars /ɡ$^ȷ$, k$^ȷ$, x$^ȷ$/, glottal /ɦ$^ȷ$/, and postalveolar /ʒ$^ȷ$, ʧ$^ȷ$, ʃ$^ȷ$, ʤ$^ȷ$/. In Ukrainian, this phenomenon occurs mainly before < і >, and thus semi-palatalized sounds are the combinatorial allophones of the respective ‘hard’ consonants. However, in a few Ukrainian words labial < в > and < м > can appear before < я > or < ьо > (*свято, духмяний, тьмяний, цвьохнути*), having a sense-distinguishing role in, e.g., *свят* /sʲʋ$^ȷ$ɑt/ (‘holiday’, Gen. Pl.) versus *сват* /sʋɑt/ (‘matchmaker’; ‘father of the son- or daughter-in-law’, Nom. Sing.) (Šerech 1951: 377). In the pronunciation of many speakers, there is a tendency to substitute semi-soft labials with a ‘labial + /j/’ combination: /ʋʲ/ → /ʋj/, /bʲ/ → /bj/, etc. (Bilodid 1969: 240), cf. also similar tendency in Polish (Swan 2002: 12). In loan-words, semi-palatal consonants, except postalveolar, can be found more frequently (*бюро, кюре, мюон, фюзеляж, ґяур*).

Semi-palatalized postalveolar /ʒʲ, ʧʲ, ʃʲ/ appear in most cases as geminate consonants in a stem-final position: *збіжжя*, *затишшя*, *ніччю*.

Semi-palatalized consonants are not found in the opposition of the respective hard consonants, except a very limited number of cases. Therefore, they are not treated as separate phonemes but as the allophones. However, it is possible that the phonological system of the Ukrainian language can change when the number of commonly used loans with semi-palatalized consonants becomes substantial.

### 2.5. On the phonemic status of geminates

In Ukrainian, geminate consonants appear mainly within morpheme boundaries. As a result of word formation, the gemination is produced by prefixation (*беззвучно*: *без* + *звучно*), suffixation (*законний: закон* + *н* + *ий*), or stem concatenation (*юннат*: *юн(ий)* + *нат(ураліст)*). In some Ukrainian words, geminates are preserved historically (*панна* 'young lady; miss', *манна* 'manna') and have a sense-distinguishing role (cf. *пана* 'gentleman; sir' Gen. Sing., *мана* 'delusion'). Another source of geminates is connected with the loss of jers in the suffix < *ьj > (Bethin 1992): *знання*, *зілля*, *життя*, *сіллю*, *ніччю*, *затишшя*, *збіжжя*, *відповіддю*, *маззю*. This produces geminate dentals /dʲ, zʲ, lʲ, nʲ, sʲ, tʲ, ʦʲ, ʣʲ/ and postalveolar /ʒʲ, ʧʲ, ʃʲ/ (the phoneme /ʤ/ is too rare to occur in this position). It is interesting that labials /b, ʋ, m, p, f/, as well as /r/, are not geminated in such situations but appear as a 'consonant + /j/' combination: *любов'ю*, *верф'ю*, *пір'я*. In all the described cases, the geminate consonants are generally treated as a sequence of two identical phonemes, not a separate phoneme (Bilodid 1969; Ponomariv 2001; Žovtobrjukh and Kulyk 1965).

### 2.6. Assimilation

In modern Ukrainian, the regressive assimilation occurs in some consonant clusters. The following types of the assimilation are known (Pohribnyj 1984; Ponomariv 2001; Žovtobrjukh and Kulyk 1965; see also Wetzels and Mascaró 2001 for comparison with some other languages).:

1) ***Regressive voicing and devoicing***

- A voiceless consonant followed by a voiced obstruent undergoes the voicing: *боротьба* /bɔrɔdʲba/, *просьба* /prɔzʲba/, *якби* /jagbɪ/, *вокзал* /ʋɔgzal/, *хоч би* /xɔʤbɪ/.
- A voiced /ɦ/ represented by < г > is devoiced when followed by a voiceless consonant: *нігті* /nʲixtʲi/, *легко* /lɛxkɔ/, *дьогтю* /dʲɔxtʲu/.
- The prefix and the preposition given by < з > is devoiced before voiceless consonants: *зсипати* /ssɪpatɪ/, *зсунути* /ssunutɪ/, *зцідити* /sʦʲidɪtɪ/, *з хати* /sxatɪ/. However, this effect is not universal and even denied by some authors (Ponomariv 2001: 18). Note, in particular, that such assimilation is reflected in orthography before < к, п, т, ф, х >: *скласти*, *спитати*. As a rule, < з > in the prefixes < роз- > and < без- > is not devoiced.

- It must be noted that voiced consonants are not devoiced when followed by voiceless: *ложка* /lɔʒka/, *казка* /kazka/, *кладка* /kladka/. Additionally, there is no final devoicing in Ukrainian: *хліб* /xlʲib/, *сад* /sad/, *низ* /nɪz/.

2) ***Assimilation by place and manner of articulation***
- Dentals before (hushing) sibilants become (hushing) sibilants: *зшити* /ʃʃɪtɪ/.
- Hushing sibilants before dentals become dentals: *дощці* /dɔsʲʦʲi/.
- The stop represented by < т > before < ч, ш > becomes /ʧ/: *коротший* /kɔrɔʧʧɪj/.
- The stop represented by < т > before < ц > becomes /ʦ/: *коритце* /kɔrɪʦʦɛ/.

3) ***Regressive palatalization***
- Dentals followed by a soft consonant are themselves palatalized: *кінський* /kʲinʲsʲkɪj/, *пісня* /pʲisʲnʲa/, *дні* /dʲnʲi/.
- Dentals represented by < с, з, ц, дз > followed by a semisoft labial are palatalized: *свято* /sʲvʲato/, *сміх* /sʲmʲix/, *цвіт* /ʦʲvʲit/, *звір* /zʲvʲir/.

## 3. Phoneme-grapheme relation

In this section we present and analyze graphemic representations of Ukrainian phonemes.

Table 1

Vowels

| Phoneme | Graphemes | Comments and examples |
|---|---|---|
| /a/ | < а ><br>< я > | *сам*<br>*яр, м'яз, зняв* |
| /ɛ/ | < е ><br>< є ><br>< и >* | *тер*<br>*твоє, мене,*<br>*мине* |
| /i/ | < і ><br>< ї > | *ліс*<br>*з'їм* |
| /ɪ/ | < и > | *сир* |
| /ɔ/ | < о > | *гора* |
| /u/ | < у ><br>< ю > | *вулик*<br>*знаю, ллю* |

* In unstressed positions only, see Sec. 2.2.

In the table below, the following abbreviations are used for certain sets of graphemes:
- < і, я, ю, є > = < IOT > ('iotated', softening a preceding consonant);
- < з, с, дз, ц, н, л, д, т > = < DEN > (dentals);
- < б, п, в, м, ф > = < LAB > (labials);
- < б, г, ґ, д, ж, з, дж, дз > = < VOB > (voiced obstruents, to distinguish from sonorants).

Table 2

Consonants

| Phoneme | Graphemes | Comments and examples |
|---|---|---|
| /b/ | < б ><br>< п > | *брат*<br>before < VOB >: *крепдешин* |
| /ʋ/ | < в ><br>< ф >* | *вага, вона*<br>before < VOB >: the root *афган…* |
| /ɦ/ | < г ><br>< хг ><br>< х >** | *гора, луг*<br>in loan-words: *бухгалтер, цейхгауз*<br>before < VOB >: *їх друг* |
| /g/ | < ґ ><br>< к > | *ґрунт*<br>before < VOB >: *якби*, *вокзал* |
| /d/ | < д ><br>< т > | *дар, рід*<br>before < VOB >: *п'ятдесят* |
| /$d^j$/ | < д ><br><br>< дь ><br>< т ><br>< ть > | followed by < IOT >: *дяк*<br>followed by soft < DEN >: *дня*<br>*відповідь*<br>before soft < VOB >: *кіт дівся*<br>before < VOB >: *боротьба* |
| /ʒ/ | < ж ><br>< з ><br>< ш > | *жир*<br>followed by < ж, ш, ч, дж >: *зжати*<br>before < VOB >: *наш друг* |
| /z/ | < з ><br>< с ><br>< ст > | *за, віз*<br>before < VOB >: *юрисдикція*<br>before < VOB >: *шістдесят* |
| /$z^j$/ | < з ><br><br><br>< зь ><br>< ж ><br>< с ><br>< сь > | followed by < IOT >: *зілля*<br>followed by soft < DEN >: *лазня*<br>followed by semi-soft < LAB >: *звір*<br>*лізь*<br>followed by soft < с, ц >: *мажся*<br>before soft < VOB >: *мус дійти*<br>before < VOB >: *просьба* |
| /j/ | < й ><br>< ї ><br>< я ><br>< ю ><br>< є > | *його, мільйон, гай*<br>In modern Ukrainian, always = /ji/: *їжак, з'їв, країна*<br>} If preceded by the apostrophe < ' >, < ь >,<br>a vowel or in a word-initial position: *я, моя, мільярд,*<br>*п'ю, б'є, знаю* |
| /k/ | < к > | *кава* |
| /l/ | < л > | *ласка* |
| /$l^j$/ | < л ><br><br>< ль > | followed by < IOT >: *люба*<br>followed by soft < DEN >: *ллє*<br>*сіль* |
| /m/ | < м > | *мама* |

| Phoneme | Graphemes | Comments and examples |
|---|---|---|
| /n/ | < н ><br>< нт > | *наш*<br>followed by < ст >: *студентство* |
| /n$^j$/ | < н ><br><br>< нь ><br>< нт > | followed by < IOT >: *няв*<br>followed by soft < DEN >: *кінський*<br>*кінь*<br>followed by the suffix < ськ >: *студентський* |
| /p/ | < п > | *пара* |
| /r/ | < р > | *рот* |
| /r$^j$/ | < р ><br>< рь > | followed by < IOT >: *ряд*<br>only before < о >: *трьох* |
| /s/ | < с ><br>< з ><br>< ст > | *сон*<br>followed by a voiceless consonant in some cases: *зсип*<br>followed by < с, н >: *шістнадцять* |
| /s$^j$/ | < с ><br><br><br>< сь ><br>< ш ><br>< ст > | followed by < IOT >: *сім*<br>followed by soft < DEN >: *слід*<br>followed by semi-soft < LAB >: *сміх, світ*<br>*колись*<br>followed by soft < с, ц >: *смієшся*<br>followed by < ськ > or soft < ц >: *роялістський, кістці* |
| /t/ | < т > | *тихо, кіт* |
| /t$^j$/ | < т ><br><br>< ть > | followed by < IOT >: *тіло, тягти*<br>followed by soft < DEN >: *новітній*<br>*ходить* |
| /f/ | < ф > | *фонтан* |
| /x/ | < х ><br>< г > | *хата*<br>followed by < к, т > in some words: *нігті*, *легко* |
| /ʦ/ | < ц ><br>< т ><br>< тс > | *цей, цнота*<br>followed by < ц >: *коритце*<br>*тсуга, спортсмен, братство* |
| /ʦ$^j$/ | < ц ><br><br><br>< ць ><br>< ч ><br>< т ><br>< ть ><br>< с > | followed by < IOT >: *цілувати*<br>followed by soft < DEN >: *міцні*<br>followed by semi-soft < LAB >: *цвіт*<br>*цього, кінець*<br>followed by soft < с, ц >: *сорочці*<br>followed by soft < ц >: *винуватця*<br>the verbal cluster <ться> corresponds to /ʦ$^j$ʦ$^j$ɑ/: *сміється*<br>in the verbal cluster < ться > |
| /ʧ/ | < ч ><br>< щ ><br>< т > | *чай*<br>= /ʃʧ/: *ще, дощ*<br>followed by < ш, ч >: *коротший, тітчин* |

| Phoneme | Graphemes | Comments and examples |
|---|---|---|
| /ʃ/ | < ш > | *шум, ваш* |
| | < щ > | = /ʃʧ/: *щока* |
| | < с > | followed by < ш >: *вирісши* |
| | < з > | followed by < ш, ч > in a word-initial position: *зшити* |
| | < ст > | followed by < ч >: *невістчин* |
| | < ч > | followed by < н >, in some words only: *ячний* |
| /ʣ/ | < дз > | *дзвонити* |
| | < ц > | followed by < VOB >: *плацдарм* |
| | < д > | followed by < с, ц, з >: *звідси* |
| /ʣʲ/ | < дз > | followed by < IOT >: *дзінь* |
| | | followed by soft < DEN > or semi-soft < LAB >: *дзвякнути* |
| | < дзь > | *ґедзь* |
| | < ц > | before soft < VOB >: *буц діда* |
| | < ць > | before soft < VOB >: *лиць багато* |
| | < д > | followed by soft < с, ц >: *одинадцять* |
| | < дь > | followed by soft < DEN >: *підводься* |
| /ʤ/ | < дж > | *бджола* |
| | < ч > | followed by < VOB >: *хоч би* |
| | < д > | followed by < ж, ш, ч >: *швидше* |

* Grapheme < ф > before < VOB > appears as [v] being the voiced counterpart of [f]. This sound is not typical for Ukrainian. Thus, it can be treated as a combinatorial allophone of /f/.

** Grapheme < х > before < VOB > appears as [ɣ] being the voiced counterpart of [x]. Such situation occurs in native Ukrainian on the word boundaries: *тих днів*. In other situations the use of [ɣ] means incorrect pronunciation and thus [ɣ] can be treated as a voiced allophone of /x/. Cf. also similar voicing in Polish (Swan 2002: 16).

These two ambiguous possibilities will not be considered as separate graphemic representations in the following analysis of the grapheme-phoneme relation.

Bernhard and Altmann (2008) proposed the Shenton–Skees-geometric distribution

$$P_x = p(1-p)^{x-1}\left[1+a\left(x-\frac{1}{p}\right)\right], \quad x = 1,2,\dots,$$

with parameters $0 < p \le 1$ and $0 \le a \le \frac{1}{1-p} - 1$ (cf. Mačutek 2008a) as a model. In the table below one finds the distribution of graphemic representations, where $x$ is the number of possibilities how a phoneme can be represented in writing, $f(x)$ is the number of phonemes with $x$ graphemic representations (i.e., 10 phonemes are represented by 1 grapheme, 12 phonemes by 2 graphemes, etc) and $NP(x)$ are expected frequencies. Our results provide another corroboration of their hypothesis.

Table 3

Fitting the Shenton–Skees-geometric distribution to graphemic representations

| phonemes | $x$ | $f(x)$ | $NP(x)$ |
|---|---|---|---|
| /ɪ, ɔ, ʋ, k, l, m, p, r, t, f/ | 1 | 10 | 10.29 |
| /ɑ, i, u, b, ɦ, g, d, lʲ, n, rʲ, tʲ, x/ | 2 | 12 | 10.99 |
| /ɛ, ʒ, z, nʲ, s, ʦ, ʧ, ʣ, ʤ/ | 3 | 9 | 7.50 |
| / dʲ, sʲ/ | 4 | 2 | 4.40 |
| /zʲ, j/ | 5 | 2 | 2.39 |
| /ʦʲ, ʃ, ʣʲ/ | 6 | 3 | 2.44 |
| | $a = 0.7105$ $\chi^2 = 1.90$<br>$p = 0.5737$ $P = 0.59$<br>$DF = 5$ | | |

It is to be noted that the Shenton–Skees-geometric distribution (see above) yields a satisfactory fit ($P = 0.52$) also in the case that the two consonant allophones discussed under Table 2 are taken into consideration.

In the following we present a study of orthographic uncertainty in Ukrainian. We only note that some other properties (economy of script system, graphemic size, graphemic load of letters, letter utility) were investigated by Bernhard and Altmann (2008), Best and Altmann (2005), Kelih (2008) and Nemcová and Altmann (2008). As the number of analyzed languages is too small to allow constructions of models, we do not examine the properties in this paper. When more languages are investigated, Ukrainian data relevant for this direction of research can be easily mined from Tables 1, 2 and 3.

The mean orthographic uncertainty $\bar{U}$ of Ukrainian phonemes defined as follows:

$$\overline{U} = \frac{1}{N}\sum_{n} f_n \log_2 n\,,$$

where $f_n$ is the number of phonemes represented by $n$ graphemes (cf. Bernhard and Altmann 2008), yields the value $\bar{U} = 1.1227$. Mean orthographic uncertainties $\overline{U_1}, \overline{U_2}$ in two languages are significantly different if

$$z = \frac{\overline{U_1} - \overline{U_2}}{\sqrt{V\left(\overline{U_1}\right) - V\left(\overline{U_2}\right)}} > 1.96,$$

$V\left(\overline{U_1}\right), V\left(\overline{U_2}\right)$ being estimation of the uncertainties variances (it holds $V\left(\overline{U}\right) = \dfrac{s^2}{0.48N\overline{x}^2}$, where $\overline{x}^2$ and $s^2$ are the sample mean and variance of the distribution of graphemic representations, cf. Table 3). The test was derived by Bernhard and Altmann (2008). For Ukrainian we obtain $\overline{x} = 2.5526$, $s^2 = 2.1420$ and $V\left(\overline{U}\right) = 0.018022$.

Table 4 below contains the comparison of mean uncertainties for the orthographies of six languages (the $z$-values are the values of the test statistics for Ukrainian compared with the language in the respective column, significant differences are highlighted in bold). The data are taken from Bernhard and Altmann (2008) for Italian, Best and Altmann (2005) for German and Swedish, Kelih (2008) for Slovene and Nemcová and Altmann (2008) for Slovak. Note that all these orthographies are based on the Latin script. Interestingly, the orthographic uncertainty of Ukrainian is significantly higher than the ones of the other two Slavic languages (Slovak and Slovene). The relatively high value for Ukrainian can be justified either by many assimilation possibilities or by a different writing system, namely Cyrillic; of course other factors cannot be excluded at this stage of research. Comparisons with other Cyrillic-based orthographies can help find an answer.

Table 4

Mean uncertainty in various writing systems

| Language | German | Italian | Slovak | Slovene | Swedish | Ukrainian |
|---|---|---|---|---|---|---|
| $\bar{U}$ | 0.965 | 0.5641 | 0.7586 | 0.7841 | 0.797 | 1.1227 |
| $z$-value | 1.00 | **3.59** | **2.10** | **2.09** | 1.75 | - |

## 4. Ukrainian version of Cyrillic: complexity and distinctivity

When talking about the script complexity, distinctivity, etc., it is to be noted that the properties of a writing system depend also on a chosen font. We apply the composition method proposed by Altmann (2004), later slightly improved by Mačutek (2008b). In this method, a point is given a measure 1, a straight line corresponds to 2, an arch not exceeding 180° corresponds to 3; a continuous connection gets the weight 1, a crisp one 2 and a crossing evaluates to 3. Evaluation can be seen in Table 5 below.

Table 5

Complexity of Cyrillic letters (font Arial)

| **letter** | **Transliteration** | **components** | **connections** | **complexity** |
|---|---|---|---|---|
| А | a | 3×2 | 3×2 | 12 |
| Б | b | 2×2+3 | 3×2 | 13 |
| В | v | 2+2×3 | 4×2 | 16 |
| Г | h | 2×2 | 2 | 6 |
| Ґ | g | 3×2 | 2×2 | 10 |
| Д | d | 5×2+3 | 6×2 | 25 |
| Е | e | 4×2 | 3×2 | 14 |
| Є | je | 2+2×3 | 1+2 | 11 |
| Ж | ž | 3×2+3×3 | 2×1+2×2+3 | 26 |
| З | z | 4×3 | 2×1+2 | 16 |
| И | y | 3×2 | 2×2 | 10 |
| І | i | 2 | — | 2 |
| Ї | ji | 2+2×1 | — | 4 |
| Й | j | 3×2+3 | 2×2 | 13 |
| К | k | 2×3+2×2 | 2×2+1 | 18 |
| Л | l | 2×2+3 | 2×2 | 11 |
| М | m | 4×2 | 3×2 | 14 |
| Н | n | 3×2 | 2×2 | 10 |
| О | o | 2×3 | 2×1 | 8 |
| П | p | 3×2 | 2×2 | 10 |
| Р | r | 2+3 | 2×2 | 9 |
| С | s | 2×3 | 1 | 7 |
| Т | t | 2×2 | 2 | 6 |
| У | u | 2+3 | 2 | 7 |
| Ф | f | 2×3+2 | 2×1+2×3 | 16 |
| Х | kh | 2×2 | 3 | 7 |
| Ц | c | 4×2 | 3×2 | 14 |
| Ч | č | 2+3 | 2 | 7 |
| Ш | š | 4×2 | 3×2 | 14 |
| Щ | šč | 5×2 | 4×2 | 18 |
| Ь | *soft sign** | 2+3 | 2×2 | 9 |
| Ю | ju | 2×2+2×3 | 2×2+2×1 | 16 |
| Я | ja | 2×2+3 | 3×2 | 13 |

* A non-phonemic character, often transliterated as < ’ > or < j >, cf., e.g., Buk and Rovenchak (2004).

Mohanty (2007) supposed that the distribution of complexities was uniform. The hypothesis was successfully tested by him for the Oriya script and by Mačutek (2008b) for the Latin and Runic scripts.

Table 6

Distribution of complexities

| **C** | $f_C$ | **C** | $f_C$ | **C** | $f_C$ | **C** | $f_C$ | **C** | $f_C$ | **C** | $f_C$ | **C** | $f_C$ | **C** | $f_C$ | **C** | $f_C$ |
|---|---|---|---|---|---|---|---|---|---|---|---|---|---|---|---|---|---|
| **2** | 1 | **5** | 0 | **8** | 1 | **11** | 2 | **14** | 4 | **17** | 0 | **20** | 0 | **23** | 0 | **26** | 1 |
| **3** | 0 | **6** | 2 | **9** | 2 | **12** | 1 | **15** | 1 | **18** | 1 | **21** | 0 | **24** | 0 | | |
| **4** | 1 | **7** | 4 | **10** | 4 | **13** | 3 | **16** | 4 | **19** | 0 | **22** | 0 | **25** | 1 | | |

We perform the run test about the mean to test the uniformity of the distribution. Denote $I$ the inventory size, $R$ the range of complexities, $\overline{C}$ the mean complexity and $\sigma_C$ the standard deviation of complexities (we only note that for Cyrillic we have $\overline{C} = 11.79$ and $\sigma_C = 5.24$). If the data are uniformly distributed, all expected frequency values are $E = \frac{I}{R+1}$. A run is a sequence of frequencies which are either all greater than $E$ or all smaller than $E$. Hence we have $E = \frac{33}{24+1} = 1.32$ and the runs [1,0,1,0, 2,4, 1, 2,4,2, 1, 3,4, 1, 4, 0,1,0,0,0,0,0,0,1,1], i.e., 9 runs. Next, denote $n = R + 1$, $n_1$ the number of frequencies smaller than $E$ and $n_2$ the number of frequencies greater than $E$ (in this case $n = 25,\ \ n_1 = 17,\ \ n_2 = 8$). The number of runs can be considered random (and, consequently, the distribution can be considered uniform) if

$$z = \frac{|r - E(r)| - 0.5}{\sigma_r} < 1.96,$$

where $r$ is the number of runs, $E(r) = 1 + \frac{2n_1 n_2}{n}$ and $\sigma_r = \sqrt{\frac{2n_1 n_2 (2n_1 n_2 - n)}{n^2 (n-1)}}$. We obtain $z = 1.13$, which means that the uniform distribution is a good model for the distribution of complexities also in this case.

Mačutek (2008b) suggested the Poisson distribution ($P_x = \frac{e^{-\lambda}\lambda^x}{x!}, \lambda > 0$) as a model for both the number of components and the number of connections. As can be seen in the following Table 7, Cyrillic is no exception, with an excellent fit for connections. For the number of components there are not enough degrees of freedom and the usual $\chi^2$ goodness of fit test cannot be used (but at least intuitively the shape of the histogram is very similar to Poisson frequencies).

Table 7

Fitting the Poisson distribution to the numbers of components and connection

| | components | connections |
|---|---|---|
| **0** | | 2 |
| **1** | 1 | 6 |
| **2** | 9 | 10 |
| **3** | 12 | 8 |
| **4** | 8 | 5 |
| **5** | 1 | 1 |
| **6** | 2 | 1 |
| | | $\lambda$=2.49<br>$\chi^2$=1.52<br>*P*=0.91<br>*DF*=5 |

A method for measuring distinctivity of letters was introduced and described in details by Antić and Altmann (2005). In short, letters are decomposed into components (i.e., points, straight lines and arches), with orientations and connection points having differentiating functions. Differences between components are assigned weights, a difference between two letters is the minimum of sums of all components differences over all possible components permutations. Some minor refinements were added by Mačutek (2008b). Differences between letters of the Cyrillic alphabet are given below.

Table 8

Differences between Cyrillic letters

| | А | Б | В | Г | Ґ | Д | Е | Є | Ж | З | И | І | Ї | Й | К | Л | М |
|---|---|---|---|---|---|---|---|---|---|---|---|---|---|---|---|---|---|
| А | 0 | 26 | 39 | 15 | 14 | 33 | 24 | 24 | 50 | 38 | 17 | 16 | 18 | 20 | 34 | 18 | 18 |
| Б | 26 | 0 | 13 | 11 | 17 | 27 | 15 | 22 | 45 | 29 | 22 | 17 | 19 | 21 | 25 | 13 | 28 |
| В | 39 | 13 | 0 | 24 | 30 | 31 | 28 | 29 | 45 | 29 | 30 | 22 | 24 | 28 | 20 | 21 | 36 |
| Г | 15 | 11 | 24 | 0 | 6 | 25 | 12 | 14 | 39 | 28 | 11 | 6 | 8 | 14 | 23 | 7 | 17 |
| Ґ | 14 | 17 | 30 | 6 | 0 | 19 | 16 | 20 | 45 | 34 | 9 | 12 | 14 | 12 | 29 | 9 | 19 |
| Д | 33 | 27 | 31 | 25 | 19 | 0 | 23 | 35 | 59 | 48 | 28 | 31 | 33 | 26 | 38 | 18 | 31 |
| Е | 24 | 15 | 28 | 12 | 16 | 23 | 0 | 26 | 51 | 40 | 21 | 18 | 20 | 24 | 31 | 19 | 24 |
| Є | 24 | 22 | 29 | 14 | 20 | 35 | 26 | 0 | 34 | 22 | 25 | 14 | 16 | 23 | 19 | 17 | 31 |
| Ж | 50 | 45 | 45 | 39 | 45 | 59 | 51 | 34 | 0 | 29 | 45 | 33 | 35 | 42 | 27 | 41 | 51 |
| З | 38 | 29 | 29 | 28 | 34 | 48 | 40 | 22 | 29 | 0 | 34 | 22 | 24 | 32 | 23 | 30 | 40 |
| И | 17 | 22 | 30 | 11 | 9 | 28 | 21 | 25 | 45 | 34 | 0 | 12 | 14 | 3 | 29 | 17 | 10 |
| І | 16 | 17 | 22 | 6 | 12 | 31 | 18 | 14 | 33 | 22 | 12 | 0 | 2 | 15 | 17 | 13 | 18 |
| Ї | 18 | 19 | 24 | 14 | 8 | 33 | 20 | 16 | 35 | 24 | 14 | 2 | 0 | 17 | 19 | 15 | 20 |
| Й | 20 | 21 | 28 | 14 | 12 | 26 | 24 | 23 | 42 | 32 | 3 | 15 | 17 | 0 | 27 | 15 | 13 |
| К | 34 | 25 | 20 | 23 | 29 | 38 | 31 | 19 | 27 | 23 | 29 | 17 | 19 | 27 | 0 | 24 | 35 |
| Л | 18 | 13 | 21 | 7 | 9 | 18 | 19 | 17 | 41 | 30 | 17 | 13 | 15 | 15 | 24 | 0 | 23 |
| М | 18 | 28 | 36 | 17 | 19 | 31 | 24 | 31 | 51 | 51 | 10 | 18 | 20 | 13 | 35 | 23 | 0 |
| Н | 14 | 17 | 30 | 10 | 8 | 27 | 16 | 20 | 45 | 34 | 17 | 12 | 14 | 20 | 25 | 13 | 23 |
| О | 28 | 25 | 25 | 18 | 24 | 24 | 30 | 14 | 30 | 20 | 24 | 12 | 14 | 21 | 19 | 20 | 30 |
| П | 14 | 17 | 30 | 6 | 4 | 19 | 16 | 20 | 45 | 34 | 13 | 12 | 14 | 16 | 29 | 9 | 15 |
| Р | 28 | 6 | 11 | 13 | 19 | 33 | 21 | 22 | 39 | 23 | 19 | 11 | 13 | 18 | 19 | 15 | 25 |
| С | 26 | 24 | 25 | 16 | 22 | 37 | 28 | 6 | 30 | 16 | 22 | 10 | 12 | 20 | 15 | 19 | 28 |
| Т | 19 | 15 | 24 | 4 | 10 | 29 | 16 | 18 | 39 | 28 | 11 | 6 | 8 | 14 | 23 | 11 | 17 |
| У | 23 | 20 | 24 | 14 | 20 | 33 | 26 | 16 | 36 | 24 | 18 | 8 | 10 | 16 | 19 | 16 | 21 |
| Ф | 39 | 35 | 35 | 28 | 34 | 48 | 40 | 26 | 28 | 34 | 34 | 22 | 24 | 31 | 29 | 30 | 40 |
| Х | 20 | 23 | 29 | 12 | 18 | 37 | 24 | 21 | 42 | 30 | 17 | 9 | 12 | 20 | 26 | 19 | 22 |
| Ц | 20 | 23 | 36 | 12 | 6 | 21 | 20 | 26 | 51 | 40 | 15 | 18 | 20 | 18 | 35 | 15 | 22 |
| Ч | 24 | 17 | 21 | 13 | 19 | 34 | 21 | 15 | 35 | 23 | 19 | 7 | 9 | 17 | 16 | 16 | 25 |
| Ш | 20 | 23 | 36 | 16 | 10 | 25 | 20 | 26 | 51 | 40 | 19 | 18 | 20 | 22 | 35 | 19 | 26 |
| Щ | 26 | 29 | 29 | 18 | 12 | 23 | 26 | 32 | 57 | 46 | 21 | 24 | 26 | 24 | 41 | 21 | 28 |
| Ь | 28 | 6 | 11 | 17 | 19 | 33 | 21 | 22 | 39 | 23 | 19 | 11 | 13 | 18 | 19 | 19 | 29 |
| Ю | 25 | 19 | 29 | 29 | 20 | 34 | 26 | 20 | 43 | 29 | 28 | 20 | 22 | 27 | 23 | 16 | 34 |
| Я | 24 | 14 | 22 | 14 | 20 | 33 | 18 | 22 | 38 | 31 | 15 | 15 | 17 | 15 | 22 | 19 | 21 |

| | Н | О | П | Р | С | Т | У | Ф | Х | Ц | Ч | Ш | Щ | Ь | Ю | Я |
|---|---|---|---|---|---|---|---|---|---|---|---|---|---|---|---|---|
| А | 14 | 28 | 14 | 28 | 26 | 19 | 23 | 39 | 20 | 20 | 24 | 20 | 26 | 28 | 25 | 24 |
| Б | 17 | 25 | 17 | 6 | 24 | 15 | 20 | 35 | 23 | 23 | 17 | 23 | 29 | 6 | 19 | 14 |
| В | 30 | 25 | 30 | 11 | 25 | 24 | 24 | 35 | 29 | 36 | 21 | 36 | 42 | 11 | 29 | 22 |
| Г | 10 | 18 | 6 | 13 | 16 | 4 | 14 | 28 | 12 | 12 | 13 | 16 | 18 | 17 | 18 | 14 |
| Ґ | 8 | 24 | 4 | 19 | 22 | 10 | 20 | 34 | 18 | 6 | 19 | 10 | 12 | 19 | 20 | 20 |
| Д | 27 | 38 | 19 | 33 | 37 | 29 | 33 | 48 | 37 | 21 | 34 | 25 | 23 | 33 | 34 | 33 |
| Е | 16 | 30 | 16 | 21 | 28 | 16 | 26 | 40 | 24 | 20 | 21 | 20 | 26 | 21 | 26 | 18 |
| Є | 20 | 14 | 20 | 22 | 6 | 18 | 16 | 26 | 21 | 26 | 15 | 26 | 32 | 22 | 20 | 22 |
| Ж | 45 | 30 | 45 | 39 | 30 | 39 | 36 | 28 | 42 | 51 | 35 | 51 | 57 | 39 | 43 | 38 |
| З | 34 | 20 | 34 | 23 | 16 | 28 | 24 | 34 | 30 | 40 | 23 | 40 | 46 | 23 | 29 | 31 |
| И | 17 | 24 | 13 | 19 | 22 | 11 | 18 | 34 | 17 | 15 | 19 | 19 | 21 | 19 | 28 | 15 |
| І | 12 | 12 | 12 | 11 | 10 | 6 | 8 | 22 | 9 | 18 | 7 | 18 | 24 | 11 | 20 | 15 |
| Ї | 14 | 14 | 14 | 13 | 12 | 8 | 10 | 24 | 12 | 20 | 9 | 20 | 26 | 13 | 22 | 17 |
| Й | 20 | 21 | 16 | 18 | 20 | 14 | 16 | 31 | 20 | 18 | 17 | 22 | 24 | 18 | 27 | 15 |
| К | 25 | 19 | 29 | 19 | 15 | 23 | 19 | 29 | 26 | 35 | 16 | 35 | 41 | 19 | 23 | 22 |
| Л | 13 | 20 | 9 | 15 | 19 | 11 | 16 | 30 | 19 | 15 | 16 | 19 | 21 | 19 | 16 | 19 |
| М | 23 | 30 | 15 | 25 | 28 | 17 | 21 | 40 | 22 | 22 | 25 | 26 | 28 | 29 | 34 | 21 |
| Н | 0 | 24 | 8 | 19 | 22 | 14 | 20 | 34 | 18 | 14 | 15 | 14 | 20 | 19 | 16 | 20 |
| О | 24 | 0 | 24 | 19 | 8 | 18 | 14 | 14 | 20 | 30 | 14 | 30 | 36 | 19 | 24 | 22 |
| П | 8 | 24 | 0 | 19 | 22 | 10 | 20 | 34 | 18 | 10 | 19 | 14 | 16 | 23 | 20 | 20 |
| Р | 19 | 19 | 19 | 0 | 18 | 13 | 14 | 29 | 20 | 25 | 11 | 27 | 31 | 4 | 21 | 15 |
| С | 22 | 8 | 22 | 18 | 0 | 16 | 13 | 22 | 18 | 28 | 11 | 28 | 34 | 18 | 22 | 19 |
| Т | 14 | 18 | 10 | 13 | 16 | 0 | 14 | 28 | 12 | 12 | 13 | 16 | 18 | 17 | 22 | 14 |
| У | 20 | 14 | 20 | 14 | 13 | 14 | 0 | 25 | 15 | 26 | 11 | 26 | 32 | 14 | 23 | 21 |
| Ф | 34 | 14 | 34 | 29 | 22 | 28 | 25 | 0 | 31 | 40 | 24 | 40 | 46 | 29 | 34 | 32 |
| Х | 18 | 20 | 18 | 20 | 18 | 12 | 15 | 31 | 0 | 24 | 16 | 24 | 30 | 20 | 26 | 20 |
| Ц | 14 | 30 | 10 | 25 | 28 | 12 | 26 | 40 | 24 | 0 | 25 | 4 | 6 | 25 | 26 | 26 |
| Ч | 15 | 14 | 19 | 11 | 11 | 13 | 11 | 24 | 16 | 25 | 0 | 25 | 31 | 11 | 18 | 14 |
| Ш | 14 | 30 | 14 | 27 | 28 | 16 | 26 | 40 | 24 | 4 | 25 | 0 | 6 | 25 | 26 | 26 |
| Щ | 20 | 36 | 16 | 31 | 34 | 18 | 32 | 46 | 30 | 6 | 31 | 6 | 0 | 31 | 32 | 32 |
| Ь | 19 | 19 | 23 | 4 | 18 | 17 | 14 | 29 | 20 | 25 | 11 | 25 | 31 | 0 | 21 | 15 |
| Ю | 16 | 24 | 20 | 21 | 22 | 22 | 23 | 34 | 26 | 26 | 18 | 26 | 32 | 21 | 0 | 23 |
| Я | 20 | 22 | 20 | 15 | 19 | 14 | 21 | 32 | 20 | 26 | 14 | 26 | 32 | 15 | 23 | 0 |

The mean distinctivity of a letter is the sum of its differences with respect to all other letters in the alphabet divided by $I-1$.

Table 9

Mean distinctivities of Cyrillic letters

| | | | | | | | | | | | | | |
|---|---|---|---|---|---|---|---|---|---|---|---|---|---|
| А | 24.84 | Д | 31.75 | И | 20.34 | Л | 18.25 | Р | 19.53 | Х | 21.94 | Ь | 20.09 |
| Б | 20.97 | Е | 24.03 | І | 15.25 | М | 26.03 | С | 20.47 | Ц | 22.72 | Ю | 24.62 |
| В | 27.53 | Є | 21.81 | Ї | 17.22 | Н | 19.69 | Т | 16.78 | Ч | 18.53 | Я | 27.16 |
| Г | 15.53 | Ж | 41.22 | Й | 20.66 | О | 22.16 | У | 19.88 | Ш | 23.91 | | |
| Ґ | 18.06 | З | 30.53 | К | 25.53 | П | 18.69 | Ф | 31.88 | Щ | 28.28 | | |

The mean distinctivity is the mean of all mean distinctivities, for Cyrillic we obtain $\overline{D} = 22.55$, cf. Antić and Altmann (2005) and Mačutek (2008b) for distinctivity analysis of Latin and Runic scripts.

Several other hypotheses, their partial corroboration or criticism, some tests and tentative models were presented in Mohanty (2007), Altmann (2008) and Mačutek (2008b). However, we postpone the analysis until more data are available.

**Acknowledgement**

J. Mačutek was supported by the research grant VEGA 1/3016/06.